\documentclass{article}
\usepackage{algorithmic}

\usepackage{hyperref}

\usepackage[accepted]{icml2012}

\usepackage{amsmath}
\usepackage{amsthm}
\usepackage{amsfonts}
\usepackage{graphicx}
\usepackage{latexsym}
\usepackage{multicol}
\usepackage{fullpage}
\usepackage[T1]{fontenc}
\usepackage{amssymb}
\usepackage{mathrsfs}
\usepackage{color}
\usepackage{algorithmic, algorithm}
\newtheorem{definition}{Definition}

\newtheorem{proposition}{Proposition}
\newtheorem{lemma}{Lemma}

\usepackage{bbm}
\usepackage{makeidx}

\makeatother

\global\long\def\rank{\operatorname*{rank}}

\global\long\def\prox{\operatorname{prox}}

\global\long\def\diag{\operatorname{diag}}

\newcommand{\RR}{\mathbb{R}}

\newcommand{\LL}{\mathcal{L}}

\newcommand{\trans}{^{\scriptscriptstyle \top}}

\newcommand{\bX}{\mathbf X}

\icmltitlerunning{Graph Prediction in a Low-Rank and Autoregressive Setting}

\begin{document} 

\twocolumn[
%\icmltitle{Simultaneously Sparse \& Low Rank Matrix Estimation}
\icmltitle{Graph Prediction in a Low-Rank and Autoregressive Setting}
% unless you've provided the [accepted] option to the icml2012
\icmlauthor{Emile Richard}{emile.richard@cmla.ens-cachan.fr}
\icmladdress{CMLA UMR CNRS 8536, ENS Cachan \& 1000mercis, France}
            
\icmlauthor{Pierre-Andr\'{e} Savalle}{pierre-andre.savalle@ecp.fr}
\icmladdress{Ecole Centrale Paris, France}
            
\icmlauthor{Nicolas Vayatis}{nicolas.vayatis@cmla.ens-cachan.fr}
\icmladdress{CMLA UMR CNRS 8536, ENS Cachan, France}

\icmlkeywords{}

\vskip 0.3in
]

\begin{abstract} 
We study the problem of prediction for evolving graph data. We formulate the problem as the minimization of a convex objective encouraging sparsity and low-rank of the solution, that reflect natural graph properties. The convex formulation allows to obtain oracle inequalities and efficient solvers. We provide empirical results for our algorithm and comparison with competing methods, and point out two open questions related to compressed sensing and algebra of low-rank and sparse matrices.

\end{abstract} 

\section{Introduction}
We study the prediction problem where the observation is a sequence of graphs adjacency matrices $(A_t)_{0 \leq t \leq T}$ and the goal is to predict $A_{T+1}$. This type of problem arises in applications such as recommender systems where, given information on purchases made by some users, one would like to predict future purchases. In this context, users and products can be modeled as nodes of a bipartite graph, while purchases or clicks are modeled as edges.
 
In functional genomics and systems biology, estimating regulatory networks in gene expression
can be performed by modeling the data as graphs and fitting predictive models is a natural way for estimating evolving networks in these contexts.

A large variety of methods for link prediction only consider predicting from a single static snapshot of the graph - this includes heuristics \cite{liben2007link, sarkar2010theoretical}, matrix factorization \cite{koren2008factorization} or probabilistic methods \cite{taskar2003link}. More recently, some works have investigated using sequences of observations of the graph to improve the prediction, such as using regression on features extracted from the graphs \cite{Richard10}, using matrix factorization \cite{koren2010collaborative}, or in some special cases probabilistic techniques. Most techniques, however, do not explicitly take into account the  inherently sparse nature of usual sequences of adjacency matrices. In this work, we extend the work of  \cite{Richard10} to address this and propose in addition a more principled way of predicting using features extracted from the sequence of graph snapshots.

We make the following assumptions about the graph sequence (represented by adjacency matrices $A_t$): 

\begin{enumerate}
\item {\bf Low-rank.} $A_t$ has low rank. This reflects the presence of highly connected groups of nodes such as communities in social networks.

\item {\bf Autoregressive linear features.} We assume given a linear map $\omega : \RR^{n \times n}\rightarrow \RR^d$ defined by a set of $n\times n$ matrices $(\Omega_i)_{1\leq i \leq d} $ \begin{equation}\label{eq:defOm}
\omega(S) = \bigg ( \langle \Omega_1 , S \rangle , \cdots ,  \langle \Omega_d , S \rangle \bigg ) 
\end{equation}
such that the vector time series $\omega(A_t)$ has an autoregressive evolution: 
$$ \omega(A_{t+1}) = W_0\trans~\omega(A_t) + N_t$$
where $W_0 \in \RR^{d \times d}$ is a sparse matrix such that $(\omega(A_t))_{t\geq0}$ is stationary. An example of linear features is degrees that is a popularity measure in social and commerce networks. \end{enumerate}
%%%%%%%%%%%%

\section{Formulation of an optimization problem}

In order to reflect the stationarity assumption on $\omega(A_t)$ we use a convex loss function $$\ell : \RR^d \times \RR^d \rightarrow \RR_+$$ to penalize the dissimilarity between two feature vectors at successive time steps. Let us introduce

  $$ \bX_{T-1} = \left (\begin{matrix} \omega(A_0)\trans \\ \omega(A_1) \trans \\ \vdots \\ \omega(A_{T-1} ) \trans \end{matrix} \right ) \in \RR^{T \times d}$$
  and 
    $$ \bX_{T} = \left (\begin{matrix} \omega(A_1) \trans \\ \omega(A_2) \trans \\ \vdots \\ \omega(A_{T} ) \trans \end{matrix} \right ) \in \RR^{T \times d}~~.$$
We also use $\ell$ to design the elementwise extension of $\ell$ to $\bX$s.  In case of quadratic loss, we consider the following $\ell_1$ penalized regression  objective : 
  
  \begin{equation*}
 J_1(W) = \frac{1}{d}\| \bX_{T-1}W-\bX_{T}\|_2^2 + \kappa~ \|W\|_1~~.
  \end{equation*}
  
To predict $A_{T+1}$, we propose a regression term penalized by the sum of $\ell_1$ and trace norm in the same fashion as in \cite{Richard12} in order to predict the future graph $A_{T+1}$ given the prediction of its features $\omega(A_T)\trans W$ should approximate $\omega(S)$ well: 
  \begin{equation*}
 J_2(S,W) = \frac{1}{d}\| {\omega}(A_T)W-\omega(S)\|_F^2 + \tau \|S\|_* + \gamma\|S\|_1
  \end{equation*}

The overall objective function we consider here is the sum of the two partial objectives $J_1$ and $J_2$, which is convex as $J_1$ and $J_2$ are both convex.

\begin{align*}
\LL &(S,W) \doteq \frac{1}{d}\|  \bX_{T-1}W - \bX_T \|_2^2 + \kappa ~\|W\|_1  \\
 &+ \frac{1}{d} \|  \omega(A_{T})\trans W - \omega(S)\trans \|_2^2 +  \tau \|S\|_* + \gamma \|S\|_1~ .
 \end{align*}
 
 Let us introduce the linear map $\Phi$ defined by 
$$ \Phi(S,W) = \bigg ( \bX_{T-1}W , ~~\omega(S)\trans- \omega(A_T)\trans W \bigg ) ~~.$$
The objective can be written as a penalized least squared regression on the joint variable $(S,W)$:
\begin{align*}\LL(S,W) = \frac{1}{d}& \|  \Phi(S,W) - \bigg( \bX_{T},0 \bigg ) \|_2^2~ \\
 &+ \gamma \|S\|_1 + \tau \|S\|_* +\|W\|_1 ~~.\end{align*}

\section{Oracle inequality}

Define $(\delta,\epsilon\trans ) =  \bigg( \bX_{T},0  \bigg )-\Phi(A_{T+1},W_0) $, {\it i.e.}, 
\[ 
\delta =  \bX_{T} -  \bX_{T-1}W_0  
\]
and
\[
\epsilon = W_0\trans \omega(A_{T}) - \omega(A_{T+1})~~.
\]

We define $M = \sum_{i=1}^d\epsilon_i \Omega_i$, where $\Omega_i$ are defined in (\ref{eq:defOm}) and let $$\Xi = \bX_{T-1}\trans \delta - \omega(A_T)\epsilon \trans~~.$$ We defined $M$ and $\Xi$ such that they verify
$$ \langle (\delta,\epsilon) , \Phi(S,W)\rangle = \langle (M,\Xi) , (S,W)\rangle $$
The following result can be proved using the tools introduced in \cite{Koltchinskii11}.
\begin{proposition}\label{prop:splrReg}
  Let $(\widehat{S}, \widehat{W})$ be the minimizers of $\LL(S,W)$ over a convex cone $\mathcal{S}\times \mathcal{W} $. Suppose that
 \begin{enumerate}
 \item for some $\mu> 0$, and for any $S_1,S_2 \in \mathcal{S}$ and $W_1,W_2 \in \mathcal{W}$,  

\begin{align*}
\frac{1}{d}\| \Phi(S_1-S_2,W_1- & W_2) \|_2^2 \\
  \geq \mu^{-2}\bigg  ( &  \| S_1 - S_2\|_F^2 +T \|W_1 - W_2\|_F^2 \bigg ) 
\end{align*}

\item  $ \tau  \geq \frac{2\alpha}{d}   \| M \|_{op}$, 
 $\gamma \geq \frac{2 (1-\alpha )}{d}\| M \|_\infty$, $\kappa \geq 2\|\Xi\|_\infty$ for any real number $\alpha \in (0;1)$;
\end{enumerate}
then 
\begin{multline}\label{eq:oracle}
\|  \widehat{S} - A_{T+1} \|_F^2 +T \|  \widehat{W} - W_0 \|_F^2\leq\\
 \mu^2 \min \bigg \{ \frac{\mu^2}{d} \bigg ( \tau  \sqrt{\rank( A_{T+1} )} \frac{ \sqrt{2}  + 1}{2}+ \gamma  \sqrt{\| A_{T+1} \|_0}  \bigg )^2 \\+\frac{\mu^2\kappa^2}{dT} \|W_0\|_0 
,~~ 2 \tau  \| A_{T+1} \|_* + 2 \gamma \|  A_{T+1}  \|_1 +2  \kappa\|W_0\|_1 \bigg \}~.
\end{multline}
\end{proposition}
The latter inequality shows how the quality of the solution is bounded by the rank and sparsity of the future graph $A_{T+1}$, and the interplay between these two prior through the parameter $\alpha$. The dependence in $T$ quantifies the improvement of the estimation in terms of the number of observations.
\section{Algorithms}

\subsection{Generalized forward-backward algorithm for minimizing $\LL$}
We use the algorithm designed in \cite{raguet2011generalized} for minimizing our objective function. Note that this algorithm outperforms the method introduced in \cite{Richard10} as it directly minimizes $\LL$ jointly in $(S,W)$ whereas the previous method first estimates $\widehat{W}$ by minimizing a functional similar to $J_1$ and then minimizes $\LL(S,\widehat{W})$. 

In addition to this we use the novel joint penalty from \cite{Richard12} that is more suited for estimating graphs.
 The proximal operator for the trace norm is given by the shrinkage operation, if $Z = U \diag (\sigma_1, \cdots, \sigma_n) V^T$ is the singular value decomposition of $Z$, 
$$ \prox_{\tau ||.||_*}(Z) = U \diag ((\sigma_i - \tau)_+)_i V^T.$$
Similarly, the proximal operator for the $\ell_1$-norm is the soft thresholding operator defined by using the entry-wise product of matrices denoted by $\circ$: 
$$\prox_{\gamma ||.||_1} = \textrm{sgn}(Z) \circ (|Z| - \gamma)_+\,.$$
The algorithm converges under very mild conditions when the step size $\theta$ is smaller than $\frac{2}{L}$, where $L$ is the operator norm of $\Phi$.

\begin{algorithm}[tbh]
   \caption{Generalized Forward-Backward to Minimize $\LL$}
   \label{alg:gfb}
\begin{algorithmic}
   \STATE Initialize $S, Z_1, Z_2, W, q = 2$
   %\FOR{$i=1$ {\bfseries to} $n_{\textrm{iterations}}$}
   \REPEAT
   \STATE Compute $(G_S,G_W) = \nabla_{S,W} \Phi (S, W)$.
   \STATE Compute $Z_1 = \prox_{q \theta \tau ||.||_{*}} (2S - Z_1 - \theta G_S)$
   \STATE Compute $Z_2 = \prox_{q \theta \gamma ||.||_{1}} (2S - Z_2 - \theta G_S)$
   \STATE Set $S = \frac{1}{q} \sum_{k=1}^q Z_k$
      \STATE Set $W = \prox_{\theta \kappa ||.||_{1}} (W - \theta G_W)$
   \UNTIL{convergence}
   \STATE {\bfseries return} $(S,W)$ minimizing $\LL$
\end{algorithmic}
\end{algorithm}

\subsection{Non-convex Factorization Method}
An alternative method to the estimation of low-rank and sparse matrices by penalizing a mixed penalty of the form $\tau\|S\|_*+\gamma\|S\|_1$ as in \cite{Richard12} is to factorize $S = UV\trans$ where $U,V \in \RR^{n \times r}$ are sparse matrices, and penalize $\gamma ( \|U\|_1 + \|V\|_1)$. The objective function to be minimized is 

\begin{align*}
\mathcal{J}&(U,V,W) \doteq \frac{1}{d}\| \bX_{T-1}W -  \bX_T  \|_F^2 + \kappa ~\|W\|_1  \\
 &+ \frac{1}{d} \| \omega(A_{T})\trans W -  \omega(UV\trans)\trans \|_2^2 + \gamma ( \|U\|_1 + \|V\|_1) \end{align*}
which is a non-convex function of the joint variable $(U,V,W)$, making the theoretical analysis more difficult. Given that the objective is convex in a neighborhood of the solution, by initializing the variables adequately, we can write an algorithm inspired by proximal gradient descent for minimizing it. 

\begin{algorithm}[tbh]
   \caption{Minimize $\mathcal{J}$}
   \label{alg:gfb}
\begin{algorithmic}
   \STATE Initialize $U,V, W$
   %\FOR{$i=1$ {\bfseries to} $n_{\textrm{iterations}}$}
   \REPEAT
   \STATE Compute $(G_U,G_V,G_W) = \nabla_{U,V,W} \Phi (UV\trans, W)$.
      \STATE Set $U = \prox_{\theta \gamma ||.||_{1}} (U - \theta G_U)$
      \STATE Set $V = \prox_{\theta \gamma ||.||_{1}} (V - \theta G_V)$
      \STATE Set $W = \prox_{\theta \kappa ||.||_{1}} (W - \theta G_W)$
   \UNTIL{convergence}
   \STATE {\bfseries return}  $(U,V,W)$ minimizing $\mathcal{J}$
\end{algorithmic}
\end{algorithm}

\section{Numerical Experiments}

\subsection{A generative model for graphs having linearly autoregressive features}\label{sec:gen}

Let $V_0 \in \RR^{n \times r}$ be a sparse matrix, $V_0 ^\dagger $ its pseudo-inverse such, that $V_0 ^\dagger V_0 =  V_0\trans V_0 ^{{\scriptscriptstyle \top} \dagger} = I_r$. Fix two sparse matrices  $W_0 \in \RR^{r \times r}$ and $U_0 \in \RR^{n \times r}$ .
Now define the sequence of matrices $(A_t)_{t \geq 0}$ for $t = 1, 2, \cdots $ by 
\[
U_t = U_{t-1}W_0 + N_t\]
and
\[
A_t = U_tV_0\trans + M_t
\]
for i.i.d sparse noise matrices $N_t$ and $M_t$, which means that for any pair of indices $(i,j)$, with high probability $(N_t)_{i,j}=0$ and $(M_t)_{i,j}=0$. \\

If we define the linear feature map $\omega(A) = A V_0^{{\scriptscriptstyle \top} \dagger} $, note that 
\begin{enumerate}
\item The sequence $\bigg (\omega(A_t) \trans \bigg )_t = \bigg (U_t + M_t V_0^{{\scriptscriptstyle \top} \dagger} \bigg )_t$ follows the linear autoregressive relation \[ \omega(A_t) \trans = \omega(A_{t-1}) \trans W_0 + N_t +  M_t V_0^{{\scriptscriptstyle \top} \dagger} ~~.\]
\item For any time index $t$, the matrix $A_t$ is close to $U_tV_0$ that has rank at most $r$
\item $A_t$ is sparse, and furthermore $U_t$ is sparse
\end{enumerate}

\subsection{Results}
We tested the presented methods on synthetic data generated as in section (\ref{sec:gen}). In our experiments the noise matrices $M_t$ and $N_t$ where built by soft-thresholding iid noise $\mathcal{N}(0,\sigma^2)$, $n=50,T=10, r=5,d=10, \sigma = .5$. After choosing the parameters $\tau,\gamma,r$ by 10-fold cross-validation, 
we compare our methods to standard baselines in link prediction \cite{liben2007link}. We use the area under the ROC curve as the measure of performance and report empirical results averaged over 10 runs. Nearest Neighbors (NN) relies on the number of common friends between each pair of nodes, which is given by $A^2$ when $A$ is the cumulative graph adjacency matrix $\widetilde{A_T} = \sum_{t=0}^TA_t$ and we denote by Shrink the low-rank approximation of $\widetilde{A_T}$. Since $V_0$ is unknown we consider the feature map $\omega(S) = SV$ where $\widetilde{A_T} = U\Sigma V\trans$ is the SVD of $\widetilde{A_T}$.

\begin{table}[hbt]
\begin{center}
\begin{tabular}{|c||c|}
\hline Method & AUC \\
\hline
NN &    0.8691 $\pm$ 0.0168 \\   
Shrink &   0.8739 $\pm$   0.0169 \\     
  $\min \LL$ &   0.9094 $\pm$  0.0176 \\     
 $\min \mathcal{J}$  &  {\bf  0.9454 $\pm$ 0.0087}\\
\hline
\end{tabular}
\caption{Performance of algorithms in terms of Area Under the ROC Curve.}
\label{tbl:fb_denoising}
\end{center}
\end{table}
\section{Discussion}
The experiments suggest the empirical superiority of the proposed approaches to the standard baselines. It is very intriguing that the non-convex matrix factorization outperforms the convex rival. A possible explanation is that minimizing the nuclear norm by using the shrinkage operator results in factorizations of the solution by two orthogonal matrices, which conflicts with the sparsity of the solution. The other benefit of the non-convex formulation is its scalability, as the proximal method proposed for the convex formulation scales in $O(n^2)$ in storage and $O(n^3)$ in time.
Several questions open perspectives for further investigations.
\begin{enumerate}

 \item  {\it Choice of the feature map $\omega$.} In the current work we used the projection onto the vector space of the top-$r$ singular vectors of the cumulative adjacency matrix as the linear map $omega$, and this choice has shown empirical superiority to other choices. The question of choosing the best measurement to summarize graph information as in compress sensing seems to have both theoretical and application potential.

\item {\it Characterization of sparse and low-rank matrices.} Can all the sparse and low-rank matrices $S$ be written as $S = UV\trans$ where $U,V\in \RR^{n \times r}$ are both sparse?  Or in other terms, what is the relation between the solution of problems penalized by $\|U\|_1+\|V\|_1$ -such as $\mathcal{J}$- and those, {\it e.g.} $\LL$, penalized by $\|S\|_1 + \beta \|S\|_*$ ?

 \end{enumerate}

  %%%%%%%%%%%%%%%%%
  %%%%%%%%%%%%%%%%%
  
  \appendix[Appendix : Proof of proposition (1)]
  
%%%%%%%%%%preuve
\section{Preliminary Tools}

\begin{definition}[Orthogonal projections associated with $S$] Let $S \in \RR^{n \times n}$ be a rank $r$ matrix.
We can write the SVD of $S$ in two ways:  $S =  \sum_{j=1}^r \sigma_j u_j v_j\trans $ or $S = U \Sigma V\trans$, where $U,V \in \RR^{n \times r}$ are orthogonal and $\Sigma = \diag(\sigma_1, \cdots , \sigma_r)$. Let $U^\perp$ and $V^{\perp}$ matrices of size $n \times (n-r)$ ortho-normally completing the bases of $U$ and $V$, and define the projections onto the vector spaces spanned by vectors $u_i$ and $v_i$ for $i = 1, \cdots r$:
   $$P_{U} = U U\trans, ~~~~~P_{U^\perp} = U^\perp U^{\perp \trans}  $$
 $$P_{V} =   V V\trans, ~~~~P_{V^{\perp}} =   V^{\perp} V^{\perp \trans} $$
and define the orthogonal projection

   $$\mathcal{P}_{S}(B) = B -   P_{U^\perp} B P_{V^\perp}    $$
  We highlight the fact that $\mathcal{P}_{S}(B)$ can also be written as

   $$\mathcal{P}_{S}(B) = P_U B P_V + P_U B P_{V^\perp} + P_{U^{\perp}} B P_V$$
   or
    $$\mathcal{P}_{S}(B) = P_U B + P_{U^{\perp}} B P_V$$

\end{definition}
We know that $\rank(\mathcal{P}_{S}(B)) \leq \rank(B)$ and $\rank(P_U B P_V ) \leq \rank(B)$. The two following inequalities will also be useful:
\begin{lemma}[Rank inequalities]
For any matrix $B$,
\begin{enumerate}
\item $\rank(\mathcal{P}_{S}(B)) \leq 2~\rank(S)$
\item $\rank(\ P_U B P_V) \leq \rank(S)$
\end{enumerate}
\end{lemma}

\begin{lemma}[Orthogonality of the decomposition]\label{lem:orth}
For any matrix $B$, with the same notations, we have
$$B =  P_{U^\perp} B P_{V^\perp}  + P_U B P_V + P_U B P_{V^\perp} + P_{U^{\perp}} B P_V$$
and the 4 terms are pairwise orthogonal. It follows that

\begin{enumerate}
\item We have the identity \begin{multline*}\|B\|_F^2 =  \|P_{U^\perp} B P_{V^\perp}\|_F^2  + \|P_U B P_V\|_F^2 \\ + \|P_U B P_{V^\perp}\|_F^2 +\| P_{U^{\perp}} B P_V\|_F^2
\end{multline*}
\item It follows that $ \|P_U B P_V\|_F \leq \|\mathcal{P}_S(B)\|_F $
\end{enumerate}
\end{lemma}

\section{Proof}

 We have  for any $(S,W) \in \mathcal{S}\times  \mathcal{W}$, by optimality of $(\widehat{S}, \widehat{W})$:
 \begin{multline}
\frac{1}{d} \bigg (\|\Phi(\widehat{S} -  A_{T+1} , \widehat{W}-W_0) \|_F^2 \\ -  \|\Phi(S -  A_{T+1} , W-W_0) \|_F^2 \bigg ) \\ = \frac{1}{d} \bigg (  \|\Phi(\widehat{S}, \widehat{W})\|_F^2 -  \|\Phi(S, W)\|_F^2 \\
 - 2 \langle \Phi(\widehat{S} - S, \widehat{W} - W) ,\Phi (A_{T+1},W_0)\rangle \bigg ) \\
 \leq \frac{2}{d} \langle \Phi(\widehat{S} - S,  \widehat{W}-W) ,  \bigg( \bX_{T},0  \bigg ) - \Phi(A_{T+1},W_0) \rangle \\
 + \tau ( \|S\|_* - \|\widehat{S}\|_* ) + \gamma ( \|S\|_1 - \|\widehat{S}\|_1 )+ \kappa  ( \|W\|_1 - \|\widehat{W}\|_1 )\\
  = \frac{2}{d} \langle (\widehat{S} - S,\widehat{W} - W)  , (M,\Xi) \rangle\\
   + \tau ( \|S\|_* - \|\widehat{S}\|_* ) + \gamma ( \|S\|_1 - \|\widehat{S}\|_1 ) + \kappa (  \|W\|_1 - \|\widehat{W}\|_1  )
 \end{multline}
Thanks to trace-duality and $\ell_1$-duality we have $ \langle M ,X \rangle \leq \|M\|_\infty \|X\|_1$ and $ \langle M ,X\rangle \leq \|M\|_{op} \|X\|_*$ for any $X$, so for any $\alpha \in [0;1]$,
 \begin{multline}\frac{1}{d} \| \Phi ( \widehat{S} - A_{T+1}, \widehat{W} - W_0) \|_F^2 \\
 \leq \frac{1}{d} \|\Phi(S - A_{T+1}, W-W_0)\|_F^2 + \tau  \|S\|_* \\
 - \tau \|\widehat{S}\|_* + 2 \alpha \| \widehat{S} - S \|_*\| M \|_{op}  \\
 + \gamma  \|S\|_1 - \gamma \|\widehat{S}\|_1 + 2 (1-\alpha) \| \widehat{S} - S \|_1\| M \|_{\infty}\\
  +  \kappa  \|W\|_1 - \kappa \|\widehat{W}\|_1 + 2  \| \widehat{W} - W \|_1\| \Xi \|_{\infty}\
 \end{multline}
now using assumptions  $ \tau  \geq \frac{2\alpha}{d}   \| M \|_{op}$, $\gamma \geq \frac{2 (1-\alpha )}{d}\| M \|_\infty$, and $\kappa \geq \frac{2}{d}\| \Xi \|_\infty$ and then triangular inequality
%\end{proof}
\begin{multline*}\frac{1}{d} \| \Phi( \widehat{S} - A_{T+1}, \widehat{W}-W_0 ) \|_F^2 \leq \\
\frac{1}{d} \| \Phi( S - A_{T+1},W-W_0 )\|_F^2 +2 \tau \|S\|_* + 2 \gamma   \|S\|_1 +2 \kappa\|W\|_1\\ \square
\end{multline*}

%%%% deuxieme partie de preuve

For proving the other bound, we start by setting some notations.
Let $S\in \mathcal{S}$, and let $r = \rank(S)$, $k = \|S\|_0$, $q = \|W\|_0$. Let $S = U \diag(\sigma_1, \cdots , \sigma_r) V\trans$ be the SVD of $S$ and let $S = \Theta_S \circ |S|$, $W = \Theta_W \circ |W|$ where $\Theta_S \in \{0, \pm 1\}^{n \times n}$, $\Theta_W \in \{0, \pm 1\}^{d \times d}$ are sign matrices of $S$,  and $W$ such that $
\| \Theta_S \|_0 = k$, $
\| \Theta_W \|_0 = q$ and $\circ$ is the entry-wise product. Let 
\begin{multline*}
Z = \tau Z_* + \gamma Z_1\\
 = \tau \bigg ( \sum_{j=1}^r u_j v_j \trans + P_{U\perp} G_* P_{V \perp} \bigg ) +  \gamma \bigg ( \Theta_S + G_1 \circ \Theta_S^\perp \bigg ) 
 \end{multline*}
denote an element of the subgradient of the convex function $S \mapsto \tau \|S\|_* + \gamma\|S\|_1$, so $\|G_*\|_{op} \leq 1$ and $\|G_1\|_\infty \leq 1$. There exist $G_1$ and $G_*$ such that

\begin{multline*} \langle Z, \widehat{S} - S \rangle = \tau \langle \sum_{j=1}^r u_j v_j \trans, \widehat{S} - S \rangle   + \tau \| P_{U\perp} \widehat{S} P_{V \perp}\|_*  \\ + \gamma  \langle \Theta_S, \widehat{S} - S \rangle + \gamma\| \Theta_S^\perp \circ \widehat{S}\|_1
 \end{multline*}

We use the two standard inequalities of convex function subdifferentials $\langle \partial \LL(\widehat{S}, \widehat{W}), (\widehat{S} - S , \widehat{W}-W)\rangle \leq 0$ and $\langle  \widehat{S} - S ,  \widehat{Z} - Z \rangle \geq 0$ and a similar inequality on subdifferentials on $\widehat{W}$ and $W$ of $W \mapsto \|W\|_1$, denoted by $\widehat{Q}$ and $Q$.  We get
 \begin{multline}\label{eq:tsybakov1}
 \langle \partial \LL(\widehat{S}, \widehat{W}),(\widehat{S}-S, \widehat{W} - W) \rangle \\ - \langle\widehat{Z} - Z,\widehat{S}-S\rangle - \langle \widehat{Q} - Q, \widehat{W} - W \rangle \leq 0
 \end{multline}
 Therefore we obtain 
 
 \begin{multline*} \langle \nabla_{(S,W)} \| \Phi(\widehat{S}, \widehat{W}) -  \bigg( \bX_{T},0  \bigg ) \|_2^2 ,(\widehat{S}-S, \widehat{W}-W) \rangle =  \\
 2\langle (\delta,\epsilon) , \Phi(\widehat{S}-S, \widehat{W}-W)  \rangle  \\
 - 2 \langle \Phi(\widehat{S}-A_{T+1}, \widehat{W}-W_0) , \Phi(\widehat{S}-S, \widehat{W}-W)  \rangle
 \end{multline*}
  The inequality (\ref{eq:tsybakov1}) can be written as
\begin{multline}
\frac{2}{d} \langle \Phi(  \widehat{S} - A_{T+1} , \widehat{W} - W_0 ), \Phi ( \widehat{S} - S, \widehat{W}-W ) \rangle \leq \\
 \frac{2}{d} \langle  (\delta,\epsilon) ,  \Phi( \widehat{S} - S, \widehat{W} - W )  \rangle \\
  -  \tau \langle \sum_{j=1}^r u_j v_j \trans, \widehat{S} - S \rangle   - \tau \| \mathcal{P}_{S}^\perp( \widehat{S})\|_*  \\ 
  - \gamma  \langle \Theta_S, \widehat{S} - S \rangle - \gamma\| \Theta_S^\perp \circ \widehat{S}\|_1\\
 - \kappa \langle \Theta_W, \widehat{W} - W \rangle - \kappa\| \Theta_W^\perp \circ \widehat{W}\|_1
\end{multline}

Thanks to Cauchy-Schwarz $$| \langle \sum_{j=1}^r u_j v_j \trans, \widehat{S} - S \rangle | \leq  \sqrt{r} \| P_{U}( \widehat{S} - S) P_{V}\|_F$$
similarly 
$$ |\langle \Theta_S, \widehat{S} - S \rangle| \leq \sqrt{k} \| \Theta_S \circ (\widehat{S} - S) \|_F$$ and $$ |\langle \Theta_W, \widehat{W} - W \rangle| \leq \sqrt{q} \| \Theta_W \circ (\widehat{W} - W) \|_F$$ 

so we have

 \begin{multline}\label{eq:Tsybakov2}
  \frac{2}{d} \langle \Phi( \widehat{S} - A_{T+1}, \widehat{W} -W_0 ) , \Phi( \widehat{S} - S,  \widehat{W} - W ) \rangle\leq \\
  \frac{2}{d} \langle (\delta,\epsilon) , \Phi( \widehat{S} - S, \widehat{W}-W ) \rangle +  \tau    \sqrt{r} \| P_{U}( \widehat{S} - S) P_{V}\|_F \\
   - \tau \| P_{U\perp} \widehat{S} P_{V \perp}\|_*  + \gamma \sqrt{k} \| \Theta_S \circ (\widehat{S} - S) \|_F - \gamma\| \Theta_S^\perp \circ \widehat{S}\|_1\\
   + \kappa \sqrt{q} \| \Theta_W \circ (\widehat{W} - W) \|_F - \kappa\| \Theta_W^\perp \circ \widehat{W}\|_1
\end{multline}

We need to bound $\langle(\delta, \epsilon) , \Phi(  \widehat{S} - S,  \widehat{W} - W ) \rangle$. For this, note that by definition, $$\langle (\delta, \epsilon) , \Phi(  \widehat{S} - S, \widehat{W} - W ) \rangle  = \langle (M, \Xi) , ( \widehat{S} - S, \widehat{W} - W  \rangle$$
and decompose for any $\alpha \in [0,1]$
\begin{multline*} M = \\
\alpha \bigg ( \mathcal{P}_{S}(M) +P_{U^\perp} M P_{V^\perp}  \bigg ) + (1 - \alpha) \bigg ( \Theta_S \circ M + \Theta_S^\perp \circ  M\bigg ) \end{multline*}
We get by applying triangle inequality, Cauchy-Schwarz, H{\"o}lder inequality written for the trace-norm and $\ell_1$-norm

\begin{multline*} \langle M ,  \widehat{S} - S  \rangle \\
\leq \alpha \bigg ( \| \mathcal{P}_{S}( M ) \|_F  \| \mathcal{P}_{S}( \widehat{S} - S )  \|_F 
\\ + \| P_{U^ \perp} M P_{V^ \perp} \|_{op}  \| P_{U^ \perp}  \widehat{S} P_{V^ \perp} \|_* \bigg ) \\
 + (1-\alpha) \bigg ( \| \Theta_S \circ M \|_F  \|  \Theta_S \circ (   \widehat{S} - S ) \|_F \\ + \|  \Theta_S^\perp \circ  M\|_{\infty}  \|  \Theta_S^\perp \circ   \widehat{S}  \|_1 \bigg )  
\end{multline*}

by rank and support inequalities obtained again by Cauchy-Schwarz
\begin{multline}\label{eq:Tsybakov3}
\langle M ,  \widehat{S} - S \rangle \\
  \leq \alpha \bigg (  \sqrt{2~r}  \| M \|_{op}  \|\mathcal{P}_{S}( \widehat{S} - S) \|_F \\ 
  +  \| M \|_{op} \|P_{U^ \perp} \widehat{S} P_{V^ \perp} \|_* \bigg ) \\
 +(1-\alpha) \bigg (  \sqrt{k} \| M \|_\infty  \| \Theta_S \circ (\widehat{S} - S) \|_F \\ 
 + \| M \|_\infty \| \Theta_S^\perp \circ \widehat{S}  \|_1   \bigg )
\end{multline}

Now by using 
\begin{multline*}
2  \langle  \Phi ( \widehat{S} - A_{T+1},  \widehat{W} - W_0 ) , \Phi ( \widehat{S} - S,  \widehat{W} - W ) \rangle = \\
\| \Phi( \widehat{S} - A_{T+1}, \widehat{W} - W_0)  \|_2^2  +  \| \Phi ( \widehat{S} - S, \widehat{W} - W )  \|_2^2  \\ -  \| \Phi ({S} - A_{T+1} , {W} - W_0)  \|_2^2 
\end{multline*}
 we can rewrite the inequality (\ref{eq:Tsybakov2}) as follows:
\begin{multline}
\frac{1}{d} \bigg ( \|   \Phi ( \widehat{S} - A_{T+1} , \widehat{W} - W_0 )   \|_2^2 \\  +  \|  \Phi (\widehat{S} - S , \widehat{W} - W ) \|_2^2  -  \|  \Phi ({S} - A_{T+1} )  \|_2^2 \bigg )   \\
\leq  \frac{2\alpha}{d} \bigg (  \sqrt{2~ r}  \| M \|_{op}  \| \mathcal{P}_{S}(\widehat{S} - S )\|_F +  \| M \|_{op} \|P_{U^ \perp} \widehat{S} P_{V^ \perp} \|_* \bigg ) \\
  +\frac{ 2(1-\alpha)}{d} \bigg (  \sqrt{k} \| M \|_\infty  \| \widehat{S} - S  \|_F + \| M \|_\infty \| \Theta_S^\perp \circ \widehat{S}  \|_1   \bigg ) \\
  +\frac{ 2}{d} \bigg (  \sqrt{q} \| \Xi \|_\infty  \| \widehat{W} -W  \|_F + \| \Xi \|_\infty \| \Theta_W^\perp \circ \widehat{W}  \|_1   \bigg )   \\
+  \tau    \sqrt{r} \|P_U(\widehat{S} - S)P_V \|_F     - \tau \| P_{U\perp} \widehat{S} P_{V \perp}\|_*  \\ 
+\gamma \sqrt{k} \| \Theta_S \circ (\widehat{S} - S) \|_F - \gamma\| \Theta_S^\perp \circ \widehat{S}\|_1 \\
 + \kappa \sqrt{q} \| \Theta_W \circ (\widehat{W} - W) \|_F - \kappa\| \Theta_W^\perp \circ \widehat{W}\|_1 \\
\leq \sqrt{r} \bigg (  \frac{2 \sqrt{2}}{d} \alpha \| M \|_{op}  + \tau \bigg ) \| \widehat{S} - S \|_F\\ + \sqrt{k} \bigg ( \frac{2(1-\alpha)}{d}  \| M \|_\infty + \gamma   \bigg )  \| \widehat{S} - S \|_F \\
+ \sqrt{q} \bigg ( \frac{2}{d}  \|\Xi \|_\infty + \kappa   \bigg )  \| \widehat{W} - W \|_F 
\end{multline}

 the last inequality being due to the assumptions $ \tau  \geq \frac{2\alpha}{d}   \| M \|_{op}$, $\gamma \geq \frac{2 (1-\alpha )}{d}\| M \|_\infty$ and $\gamma \geq \frac{2 }{d}\| \Xi \|_\infty$.

 So finally, and by using again these assumptions,

\begin{multline} \frac{1}{d} \bigg ( \|  \Phi (\widehat{S} - A_{T+1} , \widehat{W} - W_0 ) \|_2^2  +  \|  \Phi (\widehat{S} - S , \widehat{W} - W )  \|_2^2 \bigg ) \leq \\
\frac{1}{d}  \|  \Phi ( {S} - A_{T+1}) \|_2^2 +
\bigg ( \sqrt{r} \tau  (  \sqrt{2}   + 1) +2  \sqrt{k} \gamma  \bigg )  \| \widehat{S} - S \|_F \\
 + 2\sqrt{q}\kappa \|W-\widehat{W} \|_F\\
 \leq  \|  \Phi ( {S} - A_{T+1}, W-W_0 ) \|_2^2 \\ + \frac{\mu}{\sqrt{d}}
\bigg ( \sqrt{r} \tau  (  \sqrt{2}   + 1) +2  \sqrt{k} \gamma  \bigg )  \| \Phi(\widehat{S} - S , \widehat{W} - W) \|_F \\
+  \frac{2 \mu \kappa \sqrt{q}}{\sqrt{dT}}
    \| \Phi(\widehat{S} - S , \widehat{W} - W) \|_F 
\end{multline}

and $b x - x^2 \leq \left ( \frac{b}{2} \right )^2$ gives

\begin{multline*}
\frac{1}{d}  \| \Phi( \widehat{S} - A_{T+1} , \widehat{W} - W_0 )  \|_2^2 \\
\leq \frac{1}{d}  \| \Phi ( {S} - A_{T+1},  {W} - W_0  ) \|_2^2 \\
+\frac{\mu^2}{4d} \bigg ( \tau \sqrt{r}  (  \sqrt{2}   + 1) +2 \sqrt{k} \gamma  \bigg )^2 
+ \frac{ \mu^2 \kappa^2 q}{dT}
\end{multline*}

and the result follows by using 
\begin{multline*}
\frac{1}{d}\| \Phi(S_1 - S_2, W_1 - W_2) \|_2^2 
\\
\geq \mu^{-2}\| S_1 - S_2\|_F^2 + \mu^{-2}T \|W_1-W_2\|_F^2
\end{multline*}
 and setting $(S,W) = (A_{T+1},W_0)$:

\begin{multline*}   \|  \widehat{S} - A_{T+1} \|_F^2 + T\| \widehat{W} - W_0 \|_F^2 \\
 \leq \frac{\mu^4}{d}
\bigg ( \sqrt{r} \tau  (  \sqrt{2}   + 1) +2  \sqrt{k} \gamma  \bigg )^2+ \frac{ \mu^2 \kappa^2 q}{dT}~~~\square
\end{multline*}

\bibliography{graphlink4}

\bibliographystyle{icml2012}

\end{document}